\documentclass[a4paper]{svproc}

\usepackage{graphicx}
\usepackage{amsmath, amssymb}
\usepackage{tikz}
\usepackage[ruled,vlined]{algorithm2e}
\usepackage{algpseudocode}
\usepackage{subcaption}
\usepackage{booktabs}

\usepackage{url}

\usepackage[utf8]{inputenc}
\usepackage{xspace}
\usepackage{comment}
\usepackage{graphicx}
\usepackage{xcolor}
\definecolor{gg}{RGB}{0, 155, 85}
\definecolor{primarycolor}{RGB}{33,49,77}   
\definecolor{angrycolor}{RGB}{210,73,42}    

\usepackage{amsmath}
\usepackage{amssymb}
\usepackage{wrapfig}
\usepackage{amsthm}
\theoremstyle{definition}

\makeatletter
\let\NAT@parse\undefined
\makeatother
\usepackage[pdfa,colorlinks,bookmarksopen,bookmarksnumbered,allcolors=gg]{hyperref}
\usepackage{cite}

\usepackage{booktabs}
\usepackage{multirow}
\usepackage{makecell}
\usepackage{tabularx}

\usepackage[font=footnotesize]{caption}
\usepackage[font=footnotesize]{subcaption}
\usepackage[export]{adjustbox}

\usepackage[
activate   = {true},
protrusion = true,
expansion  = true,
kerning    = true,
spacing    = true,
tracking   = false,
auto       = true,
selected   = true,
factor     = 1000,
stretch    = 15,
shrink     = 15,
]{microtype}

\graphicspath{{figures/}}

\begin{document}
\mainmatter               %

\title{Asymmetric Scout--Worker Reconnaissance for Route Validation in Unknown Environments
}
\titlerunning{Asymmetric Scout--Worker Reconnaissance}

\author{
Kashif Khurshid Noori\inst{1} \and
Jaskrit Singh\inst{1} \and
Athanasios Ch. Kapoutsis\inst{2} \and
Jing Xiao\inst{1} \and
Constantinos Chamzas\inst{1}
}

\authorrunning{Noori et al.}
\tocauthor{
Kashif Khurshid Noori, Jaskrit Singh, Jing Xiao,
Athanasios Ch. Kapoutsis, Constantinos Chamzas
}

\institute{
Robotics Engineering Department,
Worcester Polytechnic Institute,
Worcester, MA, USA\\
 \email{\{kknoori,jsingh,jxiao2,chamzas\}@wpi.edu}
\and
Department of Electrical and Computer Engineering,
Democritus University of Thrace,
Xanthi, Greece\\
 \email{akapouts@ee.duth.gr}
}

\authorrunning{Noori et al.}
\tocauthor{
Kashif Khurshid Noori, Jaskrit Singh,
Athanasios Ch. Kapoutsis, Jing Xiao, Constantinos Chamzas
}
\tocauthor{Author, Co-author, Advisor}

\maketitle
\begingroup
\renewcommand{\thefootnote}{}
\footnotetext{%
 \fontsize{8}{10}\selectfont
 We acknowledge the technical and financial support of the Automotive Research Center (ARC) in accordance with Cooperative Agreement W56HZV-24-2-0001 U.S. Army DEVCOM Ground Vehicle Systems Center (GVSC) Warren, MI. DISTRIBUTION STATEMENT A. Approved for public release;
  distribution is unlimited. OPSEC11103.\par}
\endgroup

\begin{abstract}

This paper studies asymmetric scout–worker reconnaissance in unknown environments, where a small, agile autonomous scout explores routes for a larger worker robot that must visit an ordered sequence of goal locations. Because the scout has a smaller footprint and greater mobility, a scout-traversable route may be infeasible for the worker; worker feasibility must therefore be inferred from scout observations. This setting is not explicitly addressed by existing exploration and replanning methods, which typically assume a single traversability model and seek optimal paths for the same robot performing the exploration. We introduce a symbiotic scout-based framework that exploits the scout’s superior mobility to explore only the portions of the unknown environment needed to identify worker-feasible path segments connecting the ordered goals. Evaluations in simulated and real-world settings demonstrate that the proposed approach validates feasible routes, repairs blocked segments with validated worker-feasible detours, and substantially reduces scout travel compared to baseline exploration and planning methods. A real-world indoor deployment further demonstrates the scout navigating narrow corridors to identify a worker-feasible route.
\end{abstract}

\section{Introduction}
\label{sec:intro}
Autonomous robots are increasingly being deployed beyond structured roads, where large worker vehicles must traverse a terrain that is only partially known, difficult to inspect, or unsafe to enter without prior reconnaissance. In such missions, ranging from planetary and disaster-response operations to forestry and mining, 
\begin{figure}
    \centering
    \includegraphics[width=1.0\linewidth]{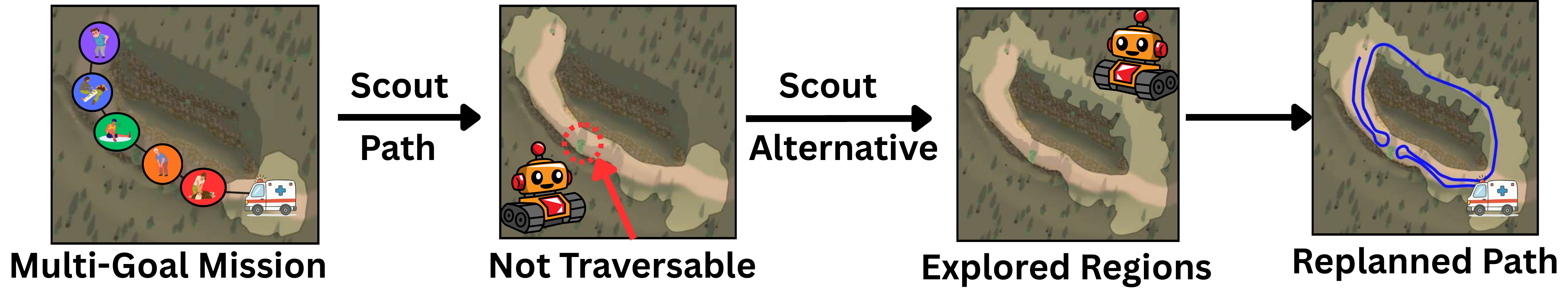}
    \caption{A scout robot explores worker-inaccessible terrain to identify feasible paths for a larger robot visiting an ordered multi-goal mission.}
    \label{fig:motivation}
    \vspace{-1.5em}
\end{figure}
reconnaissance may be performed by a smaller robot whose mobility, sensing,
footprint, and operational role differ from those of the worker vehicle. This
asymmetry creates both an opportunity and a challenge: a scout can quickly reach
viewpoints that may be inaccessible to the worker, but a route that is safe for
the scout is not necessarily traversable by the worker, as shown in \autoref{fig:motivation}. This raises the central
question: \emph{how should a scout exploit its greater mobility to move and sense so as to discover a worker-feasible path through a prescribed sequence of goal locations?}

Prior work does not directly address this asymmetric reconnaissance problem. Frontier and next-best-view planners~\cite{yamauchi1997frontier,bircher2016rhnbvp,cao2023tare,vutetakis2025apn}, task-aware exploration, and informative path planning~\cite{naazare2022wgnbvp,burusa2024semantic,schmid2020online,moon2025iatigris} generally assume that sensing, planning, and execution share a single traversability model. Heterogeneous robot teams relax the single-platform assumption, but existing methods typically map first and plan later, let each robot reason only about its own feasibility, or repeatedly replan a follower path~\cite{peterson2018aerial,kulkarni2022cohort,rockenbauer2025traversing,arora2019heterogeneous}. Incremental search methods repair paths as information arrives, but are usually single-robot, single-start--goal, and shared-collision-model formulations~\cite{koenig2002dlite,koenig2004lpa,felner2004pha,kapoutsis2017cia}. Thus, existing methods do not directly answer how a scout should gather the observations needed to identify a worker-feasible path through ordered goals.

We propose a scout-based ordered-goal route-validation framework in which exploration is guided by the worker's mobility requirements while the scout remains free to exploit its own traversability. The scout maintains a graph of observed scout-traversable space and selects reachable viewpoints based on their utility for establishing worker-feasible connectivity between consecutive goals. The method first biases exploration toward a provisional worker route through the goals, then projects the scout graph onto a worker-feasible graph. If the goal-containing components remain disconnected, a second phase directs the scout toward observations that may reveal worker-feasible detours. At termination, the framework returns a worker-feasible ordered-goal path if one is found; otherwise, it reports unresolved goal connections.

This paper makes the following contributions:
\begin{itemize}
\vspace{-0.5em}
\item A formal definition of asymmetric scout--worker route validation, where a scout with its own traversability constraints gathers information to discover routes for a larger worker with stricter mobility and feasibility constraints.
\item A two-phase scout-exploration framework for validating and repairing ordered-goal worker routes, preserving validated worker-traversable segments, inserting observed worker-feasible detours when needed, and reporting unresolved portions under the worker's footprint and traversability model.
\item Evaluations in SimpleSim and CARLA demonstrating the method's effectiveness and substantially reduced scout travel relative to baselines.
\item A real-world indoor deployment on the $2^{\text{nd}}$ floor of WPI's Robotics Dept., where the scout navigates narrow corridors to identify a worker-feasible route.
\end{itemize}
\vspace{-1.0em}
\section{Related Work}
\label{sec:related-work}
We organize related work into three areas: single-robot exploration, single-robot navigation in unknown environments, and heterogeneous scout–worker teams.

\subsection{Single Robot: \emph{Environment Exploration}}
Single robot environment exploration methods focus on planning how a robot should move in an environment to gather data about it. Early work~\cite{yamauchi1997frontier} uses the concept of frontiers, the border between known and unknown cells, to guide exploration. Later, this idea was expanded through Next-Best-Views (NBVs)~\cite{bircher2016rhnbvp}, which are possible future views for the robot. The idea is to score these NBVs based on how much information they provide, such as how many frontiers they reveal, and to choose a nearby, high-scoring view as the robot's next sub-goal. Recent graph-based methods~\cite{cao2023tare,vutetakis2025apn} reduce repeated NBV-query cost by maintaining reusable graph or hypergraph structures over the explored space. Other refinements introduce task priors into the gain function~\cite{naazare2022wgnbvp,burusa2024semantic} or formulate exploration as constrained information-cost optimization~\cite{schmid2020online,moon2025iatigris}. Across existing environment exploration planners, sensing, planning, and execution share a single robot and a single collision model, with the gain function rewarding the explorer's own progress through the environment. This differs from our setting, where a scout searches for a path for a worker with restricted mobility.

\subsection{Single Robot: \emph{Navigating to a Goal in an Unknown Environment}}

A related line of work studies goal-directed navigation in unknown environments. Dynamic replanning methods~\cite{koenig2002dlite,koenig2004lpa} repair $A^*$-style cost estimates when edge costs change, reusing prior search effort rather than restarting. The heuristic itself remains fixed. PHA*~\cite{felner2004pha} extends the setting to graphs whose topology is initially unknown, with a physical agent traversing to reveal edges and a cost metric that accumulates travel rather than node expansions. CIA*~\cite{kapoutsis2017cia} updates the heuristic at runtime based on the revealed topology. However, none of these methods consider the worker-scout formulation in this paper. Exploration and mapping assume a single robot.

\subsection{Multi-Robot: \emph{Heterogeneous Scout--Worker Teams}} 

Heterogeneous exploration systems separate sensing from execution using different sensors on one platform~\cite{arora2019heterogeneous} or physically distinct robots. Some use an aerial robot to build a map for a ground robot~\cite{peterson2018aerial}, while subterranean teams coordinate heterogeneous robots under shared coverage objectives with per-robot feasibility checks~\cite{kulkarni2022cohort}. Closest to our setting, Rockenbauer et al.~\cite{rockenbauer2025traversing} define an aerial scout's utility relative to a ground follower's optimistic path and recompute that path online. Existing methods therefore either map first and plan later, optimize each robot's own feasibility, or repeatedly replan the follower path. They do not directly address scout-guided validation of a prescribed ordered route for a worker with stricter traversability constraints.
\vspace{-0.5em}
\section{Problem Formulation}
\label{sec:prob-form}

We consider a bounded workspace $\Omega \subset \mathbb{R}^{2}$ with a known outer boundary. A robot configuration is
\begin{equation}
    q=(x,y,\theta)\in\mathcal{X},
    \qquad
    \mathcal{X}=\Omega\times\mathbb{S}^{1}.
    \label{eq:configuration_space}
\end{equation}
Configurations outside $\mathcal{X}$ are infeasible.

\looseness=-1
The system consists of a scout robot and a worker robot. We assume the worker
has no greater mobility than the scout: due to its larger footprint or stricter motion constraints, the worker's traversable set is contained in the scout's.
The two robots therefore induce different traversability maps. At online
iteration $k$, let
\begin{equation}
    T_j^k:\mathcal{X}\rightarrow
    \{\mathrm{free},\ \mathrm{obstructed},\ \mathrm{unknown}\},
    \qquad j\in\{s,w\},
    \label{eq:traversability_maps}
\end{equation}
denote the traversability map for robot $j$, where $s$ denotes the scout and
$w$ denotes the worker. For the observed portion of the workspace, this asymmetry implies
\begin{equation}
    T_w^k(q)=\mathrm{free}
    \;\Rightarrow\;
    T_s^k(q)=\mathrm{free},
    \qquad
    T_s^k(q)=\mathrm{obstructed}
    \;\Rightarrow\;
    T_w^k(q)=\mathrm{obstructed}.
    \label{eq:nested_traversability}
\end{equation}
Thus, the worker map is at least as restrictive as the scout map: a configuration
may be free for the scout but obstructed for the worker, but not the reverse.
This nested model is appropriate when feasibility differences are dominated by
footprint or clearance; strongly platform-dependent terrain interactions may
require non-nested traversability models, which we leave for future work.

The scout and worker have given safe initial configurations $q_{\mathrm{init}}^{s},q_{\mathrm{init}}^{w}\in\mathcal{X}$, which need not coincide. The worker remains at $q_{\mathrm{init}}^{w}$ while the scout explores from $q_{\mathrm{init}}^{s}$. The scout only knows what it senses at $q_{\mathrm{init}}^{s}$; subsequent map information is obtained only from measurements at visited scout configurations, and all unobserved configurations are labeled unknown. We assume a static environment and correct classification of observed configurations under each robot's traversability model.

The scout is equipped with a limited-range sensor characterized by a range $r > 0$ and an angular field of view $\phi \in (0, 2\pi]$. From a configuration $q \in \mathcal{X}$, the sensor returns measurements over the visible set $\mathrm{Vis}(q; r, \phi) \subseteq \mathcal{X}$, which contains the configurations within distance $r$ of $q$, within the angular sector of width $\phi$ centered on the scout's heading, and whose line of sight from $q$ is not blocked by an obstructed configuration. Observations outside this set are not available at $q$.

The worker task is specified only over positions. Starting from $
    g_0=\operatorname{pos}(q_{\mathrm{init}}^{w})$, $\operatorname{pos}(x,y,\theta)=(x,y),$ the worker must reach the ordered subgoal list in the prescribed order:
\begin{equation}
    \mathcal{G}=(g_1,\ldots,g_m),
    \qquad g_i\in\Omega,
    \label{eq:subgoals}
\end{equation}
 
\subsection*{Scout motion and map updates}

At iteration $k$, the scout chooses its next configuration as per an
online policy,
\begin{equation}
    q_{k+1}^{s}
    =
    \pi_k(\mathcal{I}_k)
    =
    (x_{k+1}^{s},y_{k+1}^{s},\theta_{k+1}^{s})
    \in\mathcal{X}, \; \text{with }\mathcal{I}_k=(q_k^s,T_s^k,T_w^k,\mathcal{G})
    \label{eq:decision_variable}
\end{equation}
so the decision uses only information available before the next measurement.
The continuous scout motion segment executed from $q_k^s$ to $q_{k+1}^s$ is denoted by
\begin{equation}
    \xi_k:[0,1]\rightarrow\mathcal{X},
    \qquad
    \xi_k(0)=q_k^s,
    \qquad
    \xi_k(1)=q_{k+1}^s
    \label{eq:scout_motion}
\end{equation}
The scout may move only through known scout-traversable configurations:
\begin{equation}
    T_s^k(\xi_k(\alpha))=\mathrm{free},
    \qquad
    \forall \alpha\in[0,1].
    \label{eq:scout_safety}
\end{equation}
At $q_{k+1}^s$, the scout acquires a measurement $z_{k+1}$ over the observation set
\begin{equation}
    \mathcal{O}_{k+1}
    =
    \mathrm{Vis}(q_{k+1}^s;r,\phi)
    \subseteq\mathcal{X}.
    \label{eq:visible_set}
\end{equation}
The measurement updates both traversability maps:
\begin{equation}
    (T_s^{k+1},T_w^{k+1})
    =
    \mathrm{Update}(T_s^k,T_w^k,\mathcal{O}_{k+1},z_{k+1}),
    \label{eq:map_update}
\end{equation}
where observed configurations are classified according to the corresponding
robot's traversability model and unobserved configurations retain their previous
labels: $T_j^{k+1}(q)=T_j^k(q)$, $\forall q\notin\mathcal{O}_{k+1},\quad j\in\{s,w\}$.
\subsection*{Worker-route extraction}
At iteration $k$, a worker-feasible ordered-goal path has been found if there exists a continuous path
$\tau:[0,1]\rightarrow\mathcal{X}$ and parameters
$0=t_0\leq t_1\leq\cdots\leq t_m=1$ such that
\begin{equation}
    \tau(0)=q_{\mathrm{init}}^{w},
    \qquad
    \operatorname{pos}(\tau(t_i))=g_i,\ i=1,\ldots,m,
    \qquad
    T_w^k(\tau(t))=\mathrm{free},\ \forall t\in[0,1].
    \label{eq:worker_certification}
\end{equation}
Thus, the worker may traverse a route through configurations already
labeled free in the worker traversability map.

Let $\mathcal{P}_k(q_{\mathrm{init}}^{w},\mathcal{G})$ be the set of paths
satisfying \eqref{eq:worker_certification}. If this set is nonempty, the returned worker path is the
minimum-cost ordered-goal path

\begin{equation}
    \tau_k^\star
    \in
    \arg\min_{\tau\in\mathcal{P}_k(q_{\mathrm{init}}^{w},\mathcal{G})}
    c_w(\tau),
    \label{eq:minimum_worker_route}
\end{equation}
where $c_w(\tau)$ denotes worker travel distance or cost. If $\mathcal{P}_k(q_{\mathrm{init}}^{w},\mathcal{G}) = \emptyset$, a worker-feasible ordered-goal route has not yet been validated at iteration $k$.

\subsection*{Online Exploration Planning Problem}
For a finite online scout
motion sequence $\mathbf{q}^s=(q_0^s,q_1^s,\ldots,q_K^s)$ generated by
\eqref{eq:decision_variable}, the exploration cost is
$\sum_{k=0}^{K-1} c_s(\xi_k)$, where $c_s(\xi_k)$ is the travel cost.
The online exploration planning problem is to find the lowest-cost scout motion
sequence that discovers a worker-feasible path through the ordered goals:
\begin{equation}
\begin{aligned}
    \min_{\mathbf{q}^s}
    \quad
        & \sum_{k=0}^{K-1} c_s(\xi_k) \\
    \mathrm{s.t.}
    \quad
        & q_0^s=q_{\mathrm{init}}^{s}, \\
        & \eqref{eq:scout_motion} \text{ and } \eqref{eq:scout_safety}
          \ \mathrm{hold},
          \qquad k=0,\ldots,K-1, \\
        & \mathcal{P}_K(q_{\mathrm{init}}^{w},\mathcal{G})\neq\emptyset.
\end{aligned}
\label{eq:online_exploration_planning}
\end{equation}
Here, $K$ is the terminal index of the finite scout motion sequence. The constraint $\mathcal{P}_K(q_{\mathrm{init}}^{w},\mathcal{G})\neq\emptyset$ requires that the information gathered by the scout reveals at least one worker-feasible path through all ordered goals.
The problem is online because each scout configuration in $\mathbf{q}^s$ is selected according to \eqref{eq:decision_variable} before the next measurement becomes available.
\vspace{-0.5em}
\section{Method}
\label{sec:method}

\begin{figure}[htb!]
    \centering
    \includegraphics[width=0.95\linewidth]{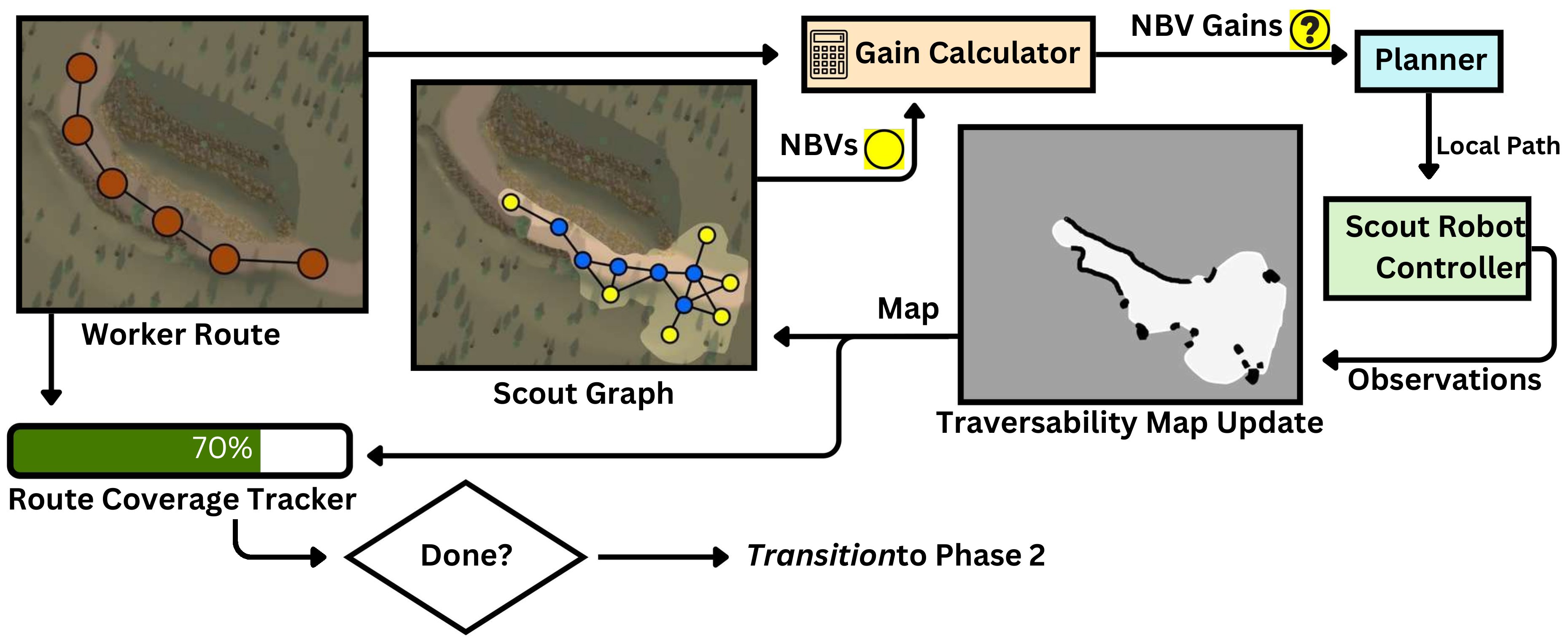}
    \caption{Phase~1: Route-biased exploration. The scout selects NBVs that reveal
    the unobserved portions of the worker corridor induced by the preferred
    route~$\sigma$.}
    \label{fig:phase-1}
    \vspace{-1.5em}
\end{figure}

\begin{figure}[t]
    \centering
    \includegraphics[width=0.95\linewidth]{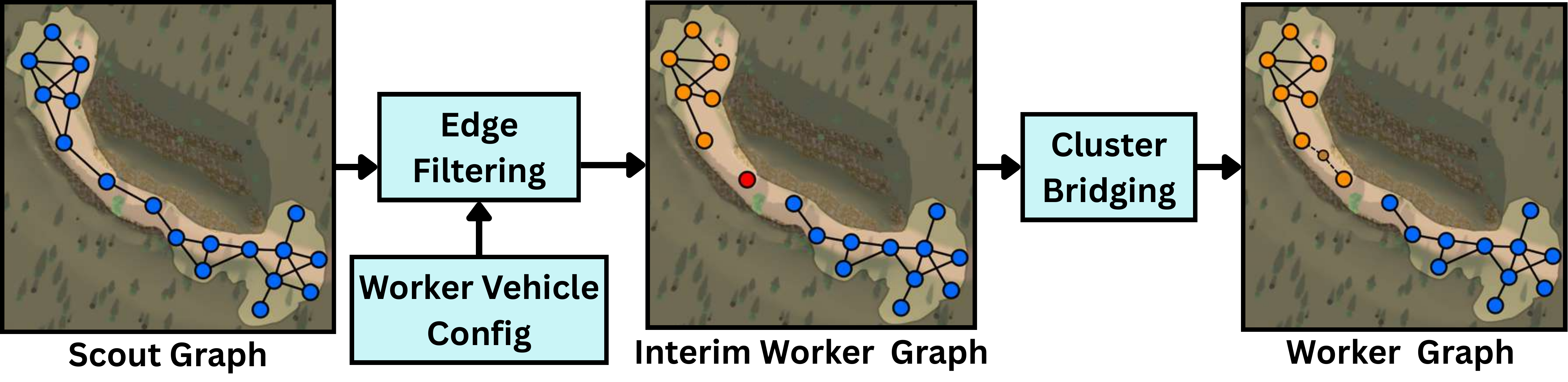}
    \caption{Transition: From the scout graph~$G_S$ to the worker graph~$G_W$}%
    \label{fig:transition}
    \vspace{-1.5em}
\end{figure}

\begin{figure}[htb!]
    \centering
    \includegraphics[width=0.85\linewidth]{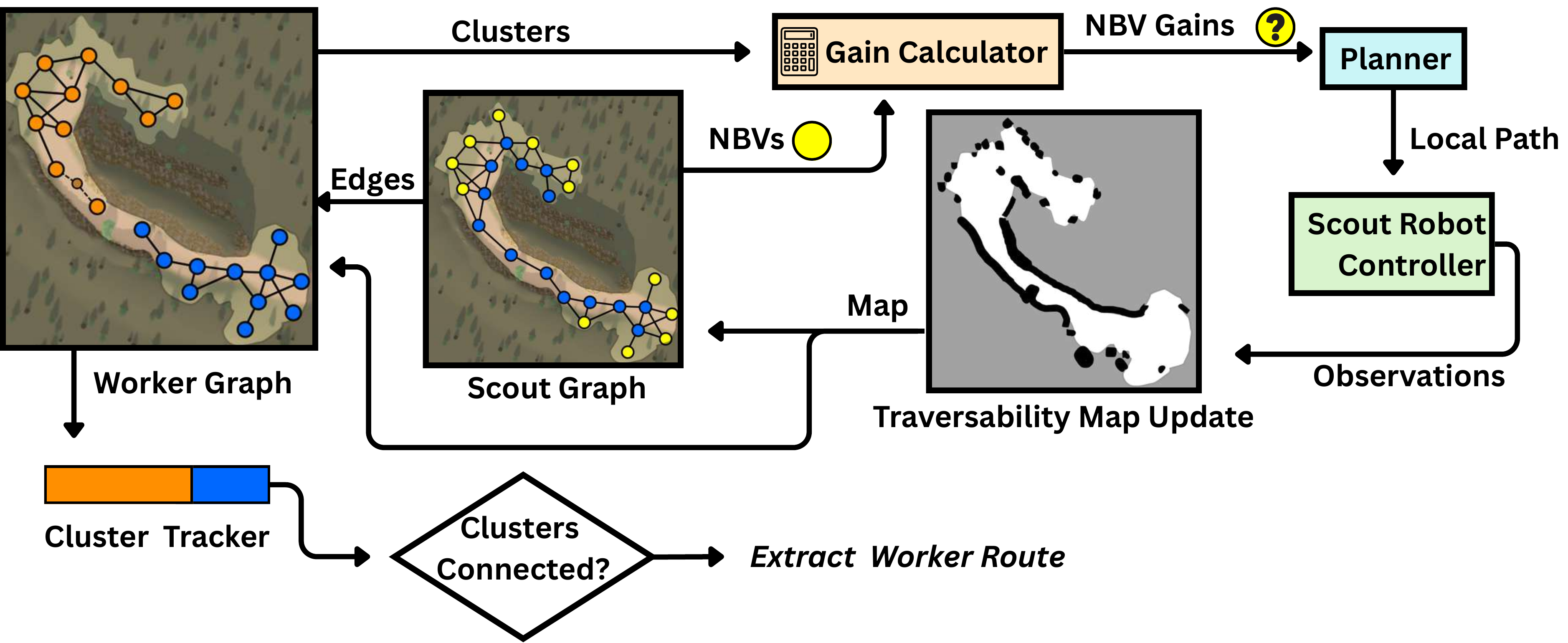}
    \caption{Phase~2: Cluster-biased detour exploration. The scout selects NBVs that may reveal worker-feasible connections between disconnected goal-containing components.}
    \label{fig:phase-2}
    \vspace{-1.5em}
\end{figure}

\looseness=-1 The method first constructs a \emph{preferred worker route}~$\sigma$ through the ordered goals. \textbf{Phase~1} attempts to validate the preferred route by
exploring its swept worker corridor. The observed scout graph is then projected to a worker-feasible graph~$G_W$. If the ordered goals are connected within a single component, the algorithm extracts a worker-feasible route; otherwise, \textbf{Phase~2} directs the scout toward observations that may reveal detours between distinct goal-containing components. Here, a connected component refers to a maximal connected subgraph of $G_W$. \autoref{alg:overview} summarizes this procedure, while \autoref{fig:phase-1}, \autoref{fig:transition}, and \autoref{fig:phase-2} illustrate the three stages.
\subsection{Preferred Worker Route}
\label{sec:route}
Given the ordered goals~$\mathcal{G}$, straight-line local connections between each pair $(g_{i-1},g_i)$ produce nominal segments $\sigma_i$, whose concatenation defines the preferred route: $\sigma = \sigma_1 \oplus \cdots \oplus \sigma_m$. This route is only a validation scaffold. \textbf{Phase~1} targets its swept worker corridor $\mathcal{R}_\sigma
    =
    \bigcup_{i=1}^{m}
    \mathrm{Sweep}_w(\sigma_i)$, whose cells must be observed to determine whether the route is worker-feasible.

\subsection{Shared Graph-Based Viewpoint Selection}
\label{sec:gains}

{
\SetAlgoSkip{}
\setlength{\interspacetitleruled}{1pt}
\setlength{\interspacealgoruled}{1pt}

\begin{algorithm}[htb!]
\caption{Overview of the scout--worker route-validation algorithm}
\label{alg:overview}
\KwIn{Ordered goals $\mathcal{G}$, map, scout and worker traversability models}
Construct preferred worker route~$\sigma$\;
\While{route-corridor coverage is below the threshold and 
$\exists$ NBV }{
    Select and execute a route-biased NBV\;
    Update the map and scout graph~$G_S$\;
}
Construct worker graph~$G_W$ from~$G_S$ and repair connectivity\;
\If{all ordered goals are connected in~$G_W$}{
    Extract a worker-feasible route and terminate\;
}
\While{goal-containing worker components remain disconnected and 
$\exists$ NBV }{
    Select and execute a cluster-biased NBV\;
    Update~$G_S$ and~$G_W$\;
}
Extract a worker-feasible route or report unresolved connectivity\;
\end{algorithm}
}
Both phases use the same graph-based viewpoint-selection model, but instantiate
it with different targets. The scout incrementally maintains a graph of observed,
scout-traversable free space and scores candidate viewpoints according to the
information they could provide for the current phase.

This graph-based abstraction follows the spirit of Next-Best-View (NBV)
exploration planners~\cite{bircher2016rhnbvp}, and specifically the graph-based variant APN~\cite{vutetakis2025apn}. The viewpoint-selection rule is inspired by
CIA*~\cite{kapoutsis2017cia}: candidate viewpoints are evaluated by trading off
their expected task progress against the cost of reaching them through the
current scout graph. Unlike generic coverage-based exploration, however, our
gains are task-aware. In \textbf{Phase~1}, they prioritize observations useful
for validating the preferred worker route; in \textbf{Phase~2}, they prioritize
observations that may reveal worker-feasible connections between disconnected
components.

\textbf{Scout Graph:}
We represent the scout's observed free space as a graph:
\begin{equation}
\label{eq:scout_graph}
G_S = (V_S, E_S),
\qquad
v = (q_v, \iota_v, \psi_v), \quad q_v \in \mathcal{X},
\end{equation}
where each node~$v\in V_S$ is a candidate or previously visited viewpoint with
configuration~$q_v$, visitation flag~$\iota_v$, and current task-aware gain
value~$\psi_v$. The current set of next-best-view candidates is
$\mathcal{N}=\{v\in V_S:\iota_v=0,\ \psi_v>0\}$. Nodes that are visited or have
zero gain are kept in~$G_S$ because they preserve reachability through the
scout's observed free space.

An edge \(e_{uv}\) enters \(E_S\) only if its local motion
\(\xi_{uv}\) satisfies the scout-safety condition in
\eqref{eq:scout_safety}. Each node also stores the frontier cells~\cite{yamauchi1997frontier} it can observe,~$\Gamma_F(v)$, and the
target cells it can observe,~$\Gamma_R(v)$. These sets provide
the geometric quantities used by the phase-dependent gain functions below.

After each scout measurement
(\autoref{eq:visible_set}--\autoref{eq:map_update}), the graph is updated only in
regions whose occupancy changed. New NBVs are sampled near newly exposed
frontiers, redundant candidates are pruned, and new edges are added. This update and pruning reduce persistence of stale frontier-driven viewpoints after map changes; no explicit frontier hysteresis or temporal smoothing is used. Both phases share this same graph and reachability structure; they differ only in how the gain field~$\psi_v$ is computed.

\textbf{Scout Planner:} The planner selects the next NBV vertex~\(v^\star\in\mathcal{N}\),
whose configuration~\(q_{v^\star}\) instantiates the next scout
decision \(q_{k+1}^s=\pi_k(\mathcal{I}_k)\) in
\eqref{eq:decision_variable}:
\begin{equation}
\label{eq:planner}
v^\star = \arg\min_{v\in\mathcal{N}} J(v),
\qquad
J(v) = \frac{d^\star(v)}{1+\eta\,\psi_v},
\end{equation}
where $d^\star(v)$ is the A*~\cite{hart1968astar} distance through~$G_S$ to
candidate~$v$, $\psi_v\geq0$ is its current task-aware gain, and $\eta\geq0$
balances travel cost against gain. Moving to~$q_{v^\star}$ is one step of the
online policy $\pi_k$ of~\eqref{eq:decision_variable}. The weighting
in~\eqref{eq:planner} mitigates local myopia by allowing a higher-gain distant
NBV to be preferred over a nearby low-gain candidate. Persistent local trapping
was not observed in our experiments, although its frequency was not quantified. The resulting online policy is heuristic and does not claim global optimality for~\eqref{eq:online_exploration_planning}.

\textbf{Next Best View Gains:} The task-aware gain~$\psi(v;\mathcal{U})$ scores
candidate vertex~$v$ by how well its observation advances an unresolved target-cell
set~$\mathcal{U}$ through three prioritized attributes.

\emph{Route Coverage} counts directly observed targets,
$\psi_C(v;\mathcal{U})=\bigl|\Gamma_R(v)\cap\mathcal{U}\bigr|$.
When none are directly observable, \emph{Frontier Visibility} rewards the
frontiers of~$v$ facing~$\mathcal{U}$, each weighted by its distance~$d_f$ to the
nearest target it faces,
$\psi_F(v;\mathcal{U})=\sum_{f\text{ facing }\mathcal{U}}\frac{1}{(1+d_f)^2}$.
Failing that, \emph{Euclidean Distance} uses proximity alone, scoring~$v$ by its
distance~$d_{\min}$ to the nearest target,
$\psi_D(v;\mathcal{U})=\frac{1}{(1+d_{\min})^2}$.
The gain takes the highest tier for which the NBV qualifies,
\begin{equation}
\label{eq:gain}
\psi(v;\mathcal{U}) =
\begin{cases}
\psi_C(v;\mathcal{U}) & \psi_C(v;\mathcal{U})>0, \\[2pt]
\beta\,\psi_F(v;\mathcal{U}) & \text{else if }\psi_F(v;\mathcal{U})>0, \\[2pt]
\gamma\,\psi_D(v;\mathcal{U}) & \text{otherwise,}
\end{cases}
\end{equation}
with \(\beta,\gamma>0\) weighting the lower gain tiers. Here, $\eta$ balances scout
travel cost against task-aware gain, while $\beta$ and $\gamma$ control their
relative weighting. For the active phase target set~$\mathcal{U}$, the stored
node gain is $\psi_v=\psi(v;\mathcal{U})$. Phase~1 uses~$\mathcal{U}_1$,
whereas Phase~2 uses~$\mathcal{U}_2(v)$; these target sets are defined in
\autoref{sec:phase1} and \autoref{sec:phase2}, respectively.

\subsection{Phase 1: Route-Biased Exploration}
\label{sec:phase1}
\begin{wrapfigure}{r}{0.32\linewidth}
    \centering
    \vspace{-6.0em}
    \includegraphics[width=\linewidth]{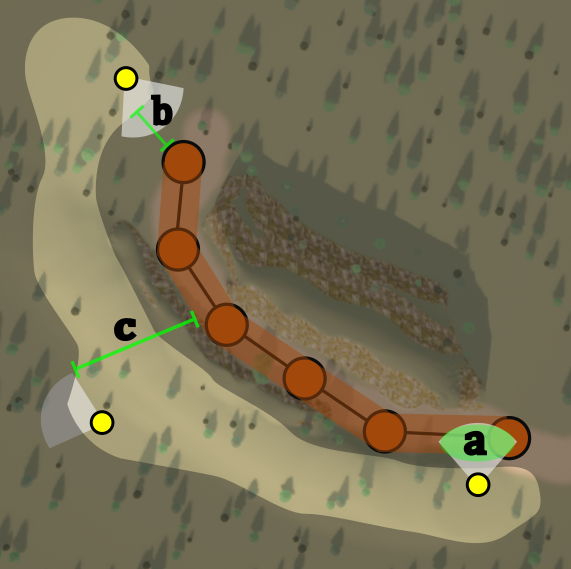}
    \caption{Phase~1 gain tiers.}
    \label{fig:gains-phase1}
    \vspace{-1.5em}
\end{wrapfigure}
Phase~1 targets the unobserved cells in the preferred worker-route corridor,
\begin{equation}
    \mathcal{U}_1
    =
    \{\, p \in \mathcal{R}_\sigma : p \text{ is unobserved} \,\},
    \label{eq:phase1_target}
\end{equation}
where \(\mathcal{R}_\sigma\) denotes the swept corridor of the worker footprint
along the route~\(\sigma\).
The scout scores each candidate~$v\in\mathcal{N}$ with the shared gain
$\psi(v;\mathcal{U}_1)$. 
As illustrated in \autoref{fig:gains-phase1}, candidate~(a) directly observes unresolved corridor cells and receives Route Coverage gain; candidate~(b) observes a frontier facing the corridor and receives Frontier Visibility gain; candidate~(c) satisfies neither condition and uses the Euclidean Distance fallback. Phase~1 terminates when the observed fraction of~\(\mathcal{R}_\sigma\) reaches the route-coverage threshold, or when no positive-gain NBV remains.

\subsection{Transition: Projecting to a Worker Graph}
\label{sec:transition}

After \textbf{Phase~1}, the scout has observed a portion of the environment and
maintains the scout graph~$G_S$. The transition step converts this graph into a
worker-feasible graph by re-evaluating $G_S$ under the
worker's stricter traversability constraints. This graph is then used to decide
whether the ordered goals are already connected for the worker, or whether
\textbf{Phase~2} must search for detours.

\paragraph{Worker graph.}
Given the current worker traversability map~$T_w^k$, we construct the worker graph \(G_W=(V_W,E_W)\) from the scout graph by retaining only configurations and local motions labeled free for the worker. In particular:
\begin{equation}
\label{eq:worker_vertices}
    V_W =
    \{\, v\in V_S : T_w^k(q_v)=\mathrm{free} \,\}
    \cup V_G ,
\end{equation}
where \(V_G=\{v_0^g,\ldots,v_m^g\}\) are explicit goal vertices associated with
the ordered goals $\mathcal{G}$, including the worker start
\(g_0=\mathrm{pos}(q^w_{\mathrm{init}})\). The worker edges are
\begin{equation}
\label{eq:worker_edges}
E_W =
\{\, e_{uv}\in E_S :
    \mathrm{Sweep}_w(e_{uv}) \subseteq \{q:T_w^k(q)=\mathrm{free}\}
\,\},
\end{equation}
where \(\mathrm{Sweep}_w(e_{uv})\) denotes the swept volume of the worker
footprint along the straight-line connection represented by~\(e_{uv}\). Each goal
vertex is attached to nearby worker-feasible vertices using the same check,
accepting only connections whose swept volume is free in~\(T_w^k\).

\paragraph{Connectivity repair.}

Because \(G_W\) is induced from sampled NBV vertices, disconnected components may reflect missing samples rather than true worker blockage. We therefore bridge nearby components by running a worker local planner, implemented here using A*~\cite{hart1968astar}, on the observed map inflated by the worker footprint, inserting successful connections as worker edges. 
Components are maintained with union-find~\cite{tarjan1975unionfind}. If one worker-feasible component contains all ordered goals, \textbf{Phase~2} is skipped; otherwise, the remaining goal-containing components are passed to \textbf{Phase~2}.

\subsection{Phase~2: Cluster-Biased Detour Exploration}
\label{sec:phase2}
If the transition step cannot validate a worker-feasible route through all ordered
goals, Phase~2 searches for detours. Let
\(\mathcal{H}=\{H_1,\ldots,H_\ell\}\) denote the goal-containing connected
components of the current worker graph~\(G_W\), where \(\ell\) is the number of
such components and each \(H_j\subseteq V_W\) is the vertex set of one component. The objective is to direct the
scout toward observations that may reveal worker-feasible connections between
these disconnected components.
For each candidate~\(v\in\mathcal{N}\), let \(H(v)\in\mathcal{H}\) denote its
home component: the component containing~\(v\) when \(v\in V_W\), or otherwise the goal-containing component nearest to~\(q_v\) in the observed map. The
Phase~2 target is candidate-dependent,
\begin{wrapfigure}{r}{0.34\linewidth}
    \centering
    \vspace{-1.0em}
    \includegraphics[width=\linewidth]{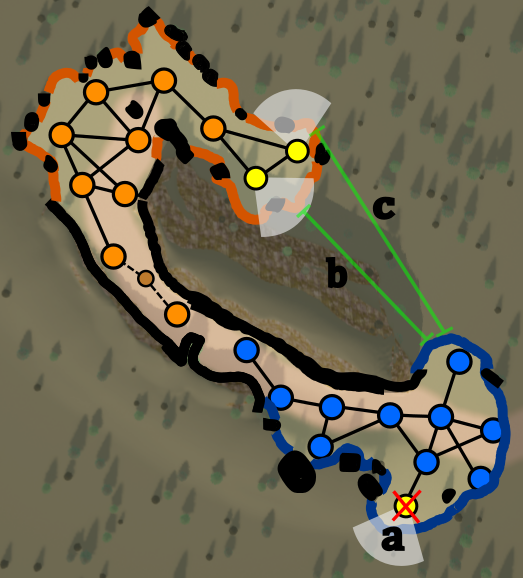}
    \caption{Phase~2 Gain tiers.}
    \label{fig:gains-phase2}
    \vspace{-1.0em}
\end{wrapfigure}
\begin{equation}
    \mathcal{U}_2(v)
    =
    \bigcup_{H_j\in\mathcal{H}\setminus\{H(v)\}}
    \mathcal{F}(H_j),
    \label{eq:phase2_target}
\end{equation}

where \(\mathcal{F}(H_j)\) denotes frontier cells adjacent to, or visible from,
component~\(H_j\). These frontiers act as proxies for unknown regions that may
connect currently disconnected worker-feasible components. The scout scores each candidate using the shared gain
\(\psi(v;\mathcal{U}_2(v))\) from \autoref{sec:gains}. 
As illustrated in
\autoref{fig:gains-phase2}, orange and blue nodes denote disconnected
worker-feasible components of~\(G_W\), while yellow points denote candidate
NBVs. Since \(\mathcal{U}_2(v)\) contains frontier cells rather than route-corridor
cells, the Route Coverage tier is inactive in Phase~2. Candidate~(b) observes
frontiers facing another component and is therefore scored by Frontier
Visibility. Candidate~(c) has no such frontier visibility and uses the Euclidean
Distance fallback. Candidate~(a) is on the blue component, which is not active for the
connection search, and receives zero gain.

After each scout observation, the worker graph~\(G_W\) is recomputed under the updated worker traversability map~\(T_w^k\), the bridging step from \autoref{sec:transition} is repeated, and the connected components are updated. Phase~2 terminates when all ordered goals lie in a single worker-feasible connected component or when no positive-gain NBVs remain. The ready-to-execute worker path can then be extracted by solving \eqref{eq:minimum_worker_route}; if $\mathcal{P}_k(q_{\mathrm{init}}^{w},\mathcal{G})=\emptyset$, no worker-feasible ordered-goal path has been validated on the current map.
\vspace{-0.5em}
\section{Experiments}
\label{sec:experiments}
We evaluate the framework along two axes: scout viewpoint-selection efficiency for candidate worker-route validation, and the full pipeline's ability to exploit scout--worker asymmetry to discover worker-feasible detours when the nominal route is blocked. Since, to the best of our knowledge, no prior method directly addresses asymmetric reconnaissance for ordered-goal worker-route validation, there is no direct end-to-end baseline; we therefore use baselines that isolate the main design choices of our approach.

Specifically, in \autoref{subsec:planner-comparison} we compare the Phase~1 task-aware viewpoint-selection policy
against two planner variants, NSGA-II and Greedy, while keeping the same
route-biased gain function fixed. This experiment isolates the effect of our
cost-aware NBV selection rule. In
\autoref{subsec:end-to-end-comparison} we evaluate the complete two-phase framework
against two end-to-end baselines: a full-exploration baseline based on
APN~\cite{vutetakis2025apn}, and a \emph{worker-as-scout} baseline in which the
exploring robot is assigned the worker footprint. The latter emulates an approach
that does not exploit scout--worker asymmetry, while the former measures the
cost of exploring broadly instead of focusing only on observations needed for
worker-route validation.

\subsection{Simulation Setup}
\label{subsec:setup}
We evaluate in two simulation environments. \textbf{SimpleSim} is a lightweight 2D occupancy-grid simulator with a \(360^\circ\) lidar, enabling controlled variation of route feasibility, detour structure, and scout start pose while keeping scout and worker traversability explicit. \textbf{CARLA}~\cite{Dosovitskiy17} provides realistic vehicle dynamics and depth sensing; there, the scout is a car-like vehicle with a front-facing \(90^\circ\) depth camera, and the worker is modeled by a larger collision footprint for feasibility checking. %
\subsection{Parameters and Metrics}
\label{subsec:params-metrics}
 
The gain parameters were selected empirically as $\eta=5$, $\beta=1$, and
$\gamma=0.7$. This configuration provided consistent behavior across the
reported experimental settings and was used throughout; no separate sensitivity
sweep was performed. Phase~1 stops at a route-coverage fraction of $0.98$.
The simulator-specific settings
follow, given as SimpleSim then CARLA. The scout is a disk of radius
$0.3$\,m, then a $4.85 \times 2.2$\,m vehicle; the worker footprint is
$3.0 \times 2.0$\,m, then $6.0 \times 2.5$\,m; the scout senses through a
$360^\circ$ $5$\,m 2D lidar, then a $90^\circ$ depth camera; the occupancy
map resolution is $0.2$\,m, then $0.3$\,m; and each configuration is run
$10$ times, then $5$. We report the total scout distance at termination,
$D_{\text{term}}$.

\subsection{Route-Coverage Experiment: Phase~1 Only}
\label{subsec:planner-comparison}

\begin{figure}[t]
\centering
\begin{subfigure}[b]{0.3\linewidth}
  \centering
  \includegraphics[width=\linewidth]{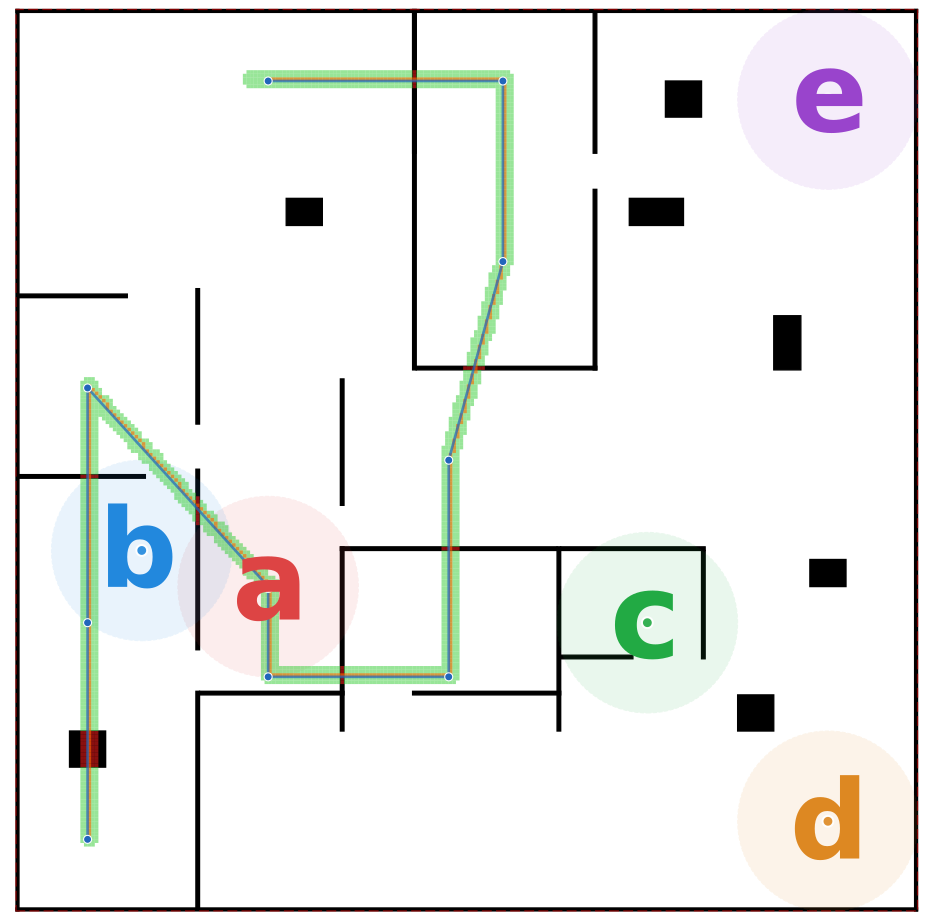}
  \caption*{Multi-Start scenario.}
  \label{fig:planner_env}
\end{subfigure}
\hfill
\begin{subfigure}[b]{0.65\linewidth}
  \centering
  \includegraphics[width=\linewidth]{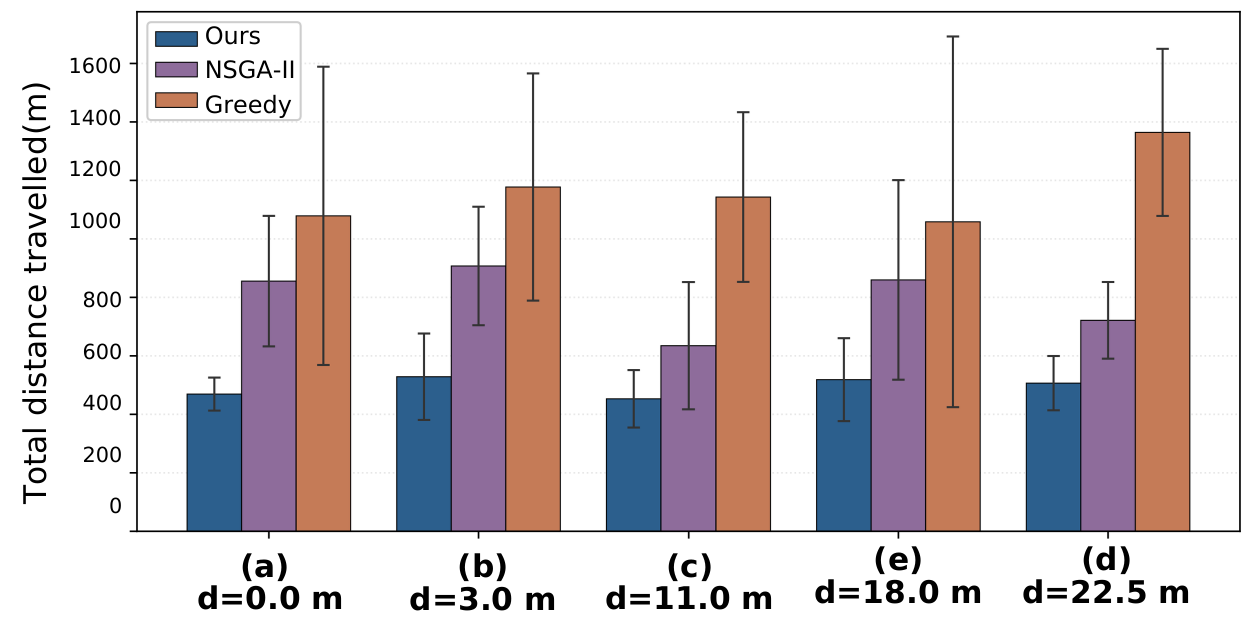}
  \caption*{Total scout travel distance by start pose.}
  \label{fig:planner_distance}
\end{subfigure}

\caption{Phase~1 route-coverage experiment. Left: five scout start poses relative to a fixed worker route. Right: total scout distance~$D_{\mathrm{term}}$ for \textit{Ours}, \textit{NSGA-II}, and \textit{Greedy}.}
\label{fig:planner_results}
\vspace{-1.5em}
\end{figure}

We first isolate the Phase~1 route-coverage stage. In this
experiment, the nominal worker route is fully worker-traversable and is held fixed,
while the scout starts from five poses at increasing distances from the route. Since the route and gain function are unchanged across
methods, differences in scout travel primarily reflect the NBV selection rule.

We compare our planner against two alternatives that use the same route-biased
gain function but differ in how they select the next NBV. The first is
NSGA-II~\cite{deb2002nsga2}, a Pareto-based optimizer implemented with
DEAP~\cite{DEAP_JMLR2012}, which solves a fixed-end open traveling-salesman
problem over the reachable NBVs with two objectives: maximizing accumulated
gain and minimizing scout travel distance. The second is a Greedy baseline that
selects the reachable NBV with the largest immediate gain. Thus, this experiment
isolates the effect of our cost-aware selection rule from the effect of the
task-aware gain itself.

Pooled across the five scout start poses, our planner terminates after
\(489.8 \pm 120.8\)\,m of scout travel, compared with
\(795.8 \pm 256.9\)\,m for NSGA-II and \(1157.8 \pm 457.9\)\,m for Greedy.
This corresponds to a \(38\%\) reduction relative to NSGA-II and a \(58\%\)
reduction relative to Greedy (\autoref{fig:planner_results}).

\subsection{End-to-End Framework Comparison}
\label{subsec:end-to-end-comparison}

We next evaluate the full two-phase framework, including route validation,
worker-graph construction, detour exploration, and final worker-route extraction.
This experiment tests whether scout--worker asymmetry and targeted detour
exploration reduce scout travel. 

\begin{figure}[t!]
\centering

\begin{subfigure}[b]{0.18\linewidth}
  \centering
  \includegraphics[width=\linewidth]{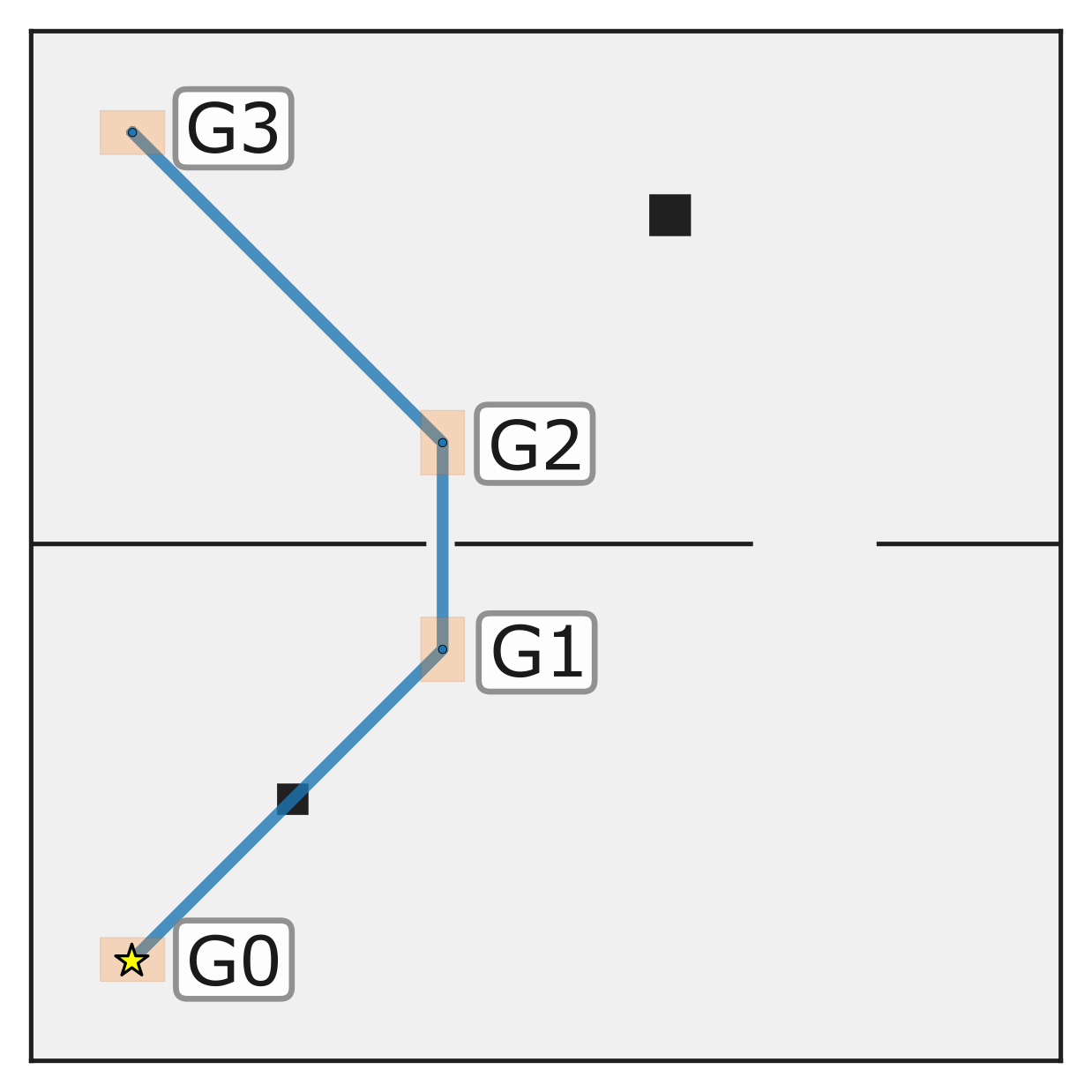}
  \caption{}
  \label{fig:env_single}
\end{subfigure}%
\hspace{0.01\linewidth}%
\begin{subfigure}[b]{0.18\linewidth}
  \centering
  \includegraphics[width=\linewidth]{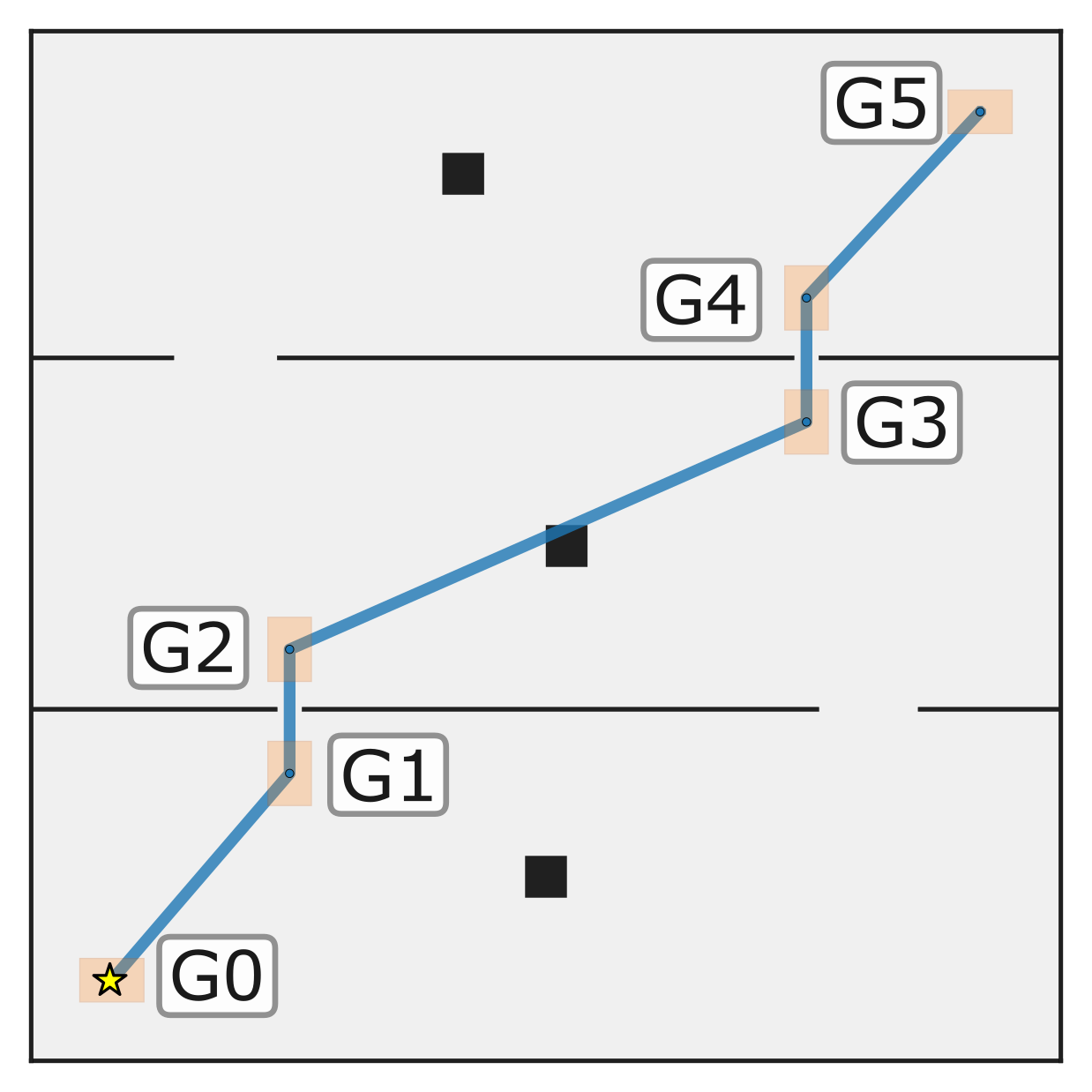}
  \caption{}
  \label{fig:env_double}
\end{subfigure}%
\hspace{0.01\linewidth}%
\begin{subfigure}[b]{0.18\linewidth}
  \centering
  \includegraphics[width=\linewidth]{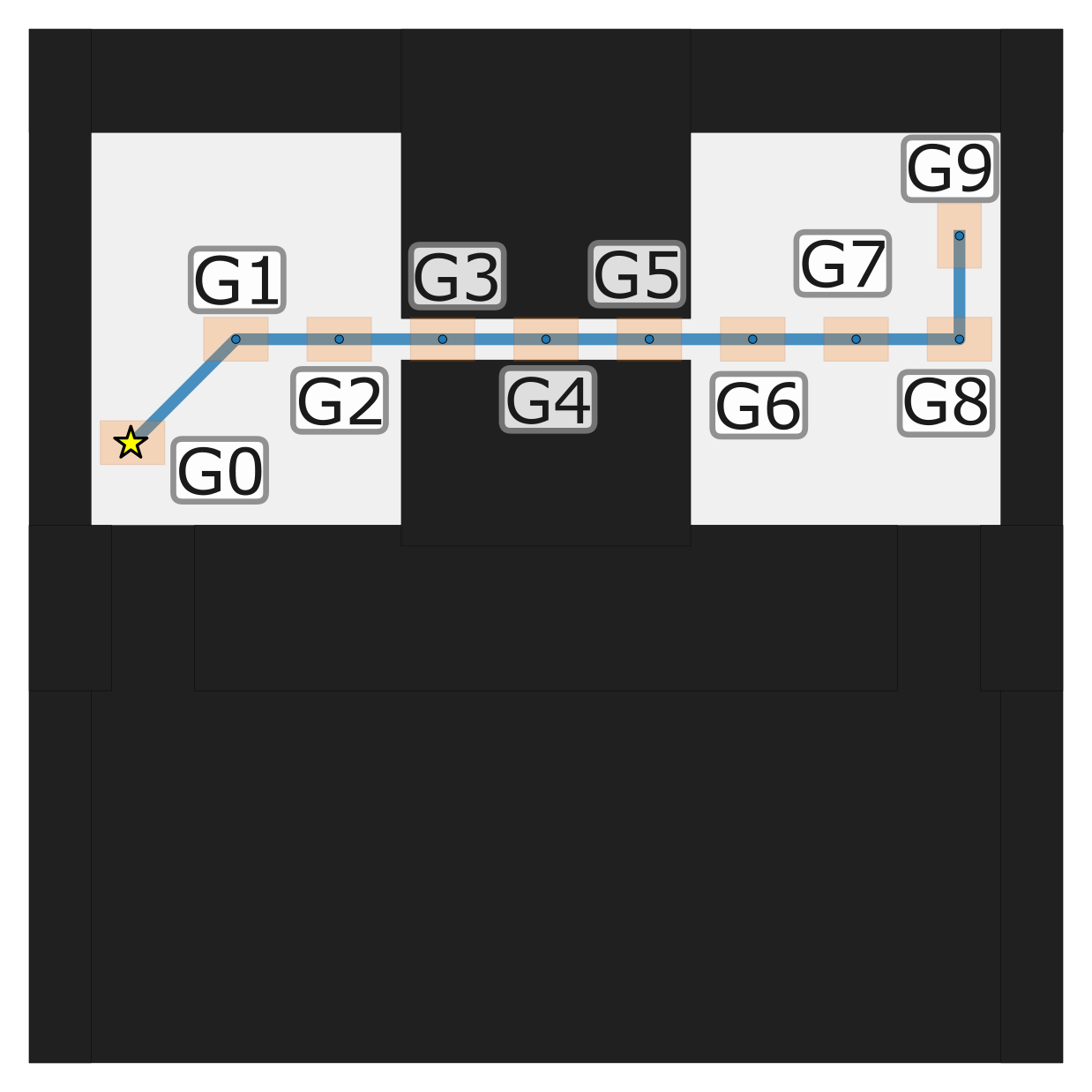}
  \caption{}
  \label{fig:env_infeasible}
\end{subfigure}%
\hspace{0.01\linewidth}%
\begin{subfigure}[b]{0.43\linewidth}
  \centering
  \includegraphics[width=\linewidth]{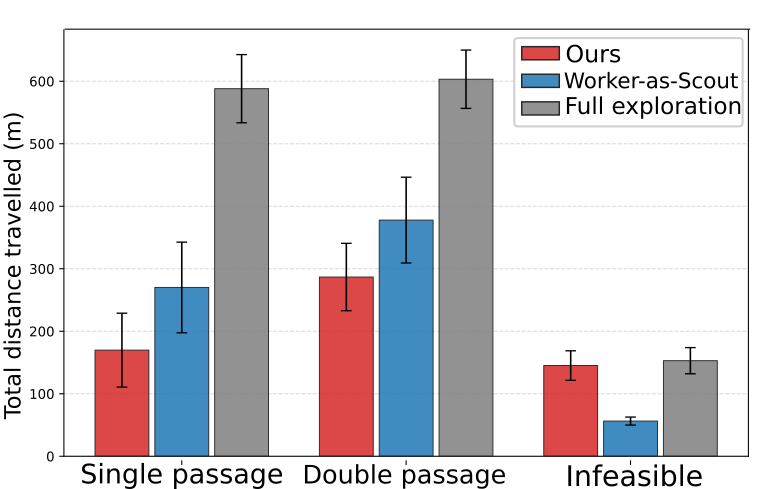}
  \caption{}
  \label{fig:scenario_distance}
\end{subfigure}

\caption{End-to-end SimpleSim results. (a)--(b) Single \& Double passage, (c) Infeasible scenario, (d) Total scout distance~$D_{\mathrm{term}}$
for \textit{Ours}, \textit{Worker-as-scout}, and \textit{Full exploration}.}
\vspace{-1.5em}
\label{fig:scenario_results}
\end{figure}

We compare against two end-to-end baselines. The \textit{Worker-as-scout}
baseline removes the asymmetry by assigning the scout the worker footprint and
running the route-validation phase with the same sensing objective. This baseline
tests whether the scout's smaller footprint and the Phase~2 repair mechanism are
actually useful. The \textit{Full exploration} baseline uses an APN-style
coverage-oriented exploration objective~\cite{vutetakis2025apn} and measures the
distance required to explore broadly rather than focusing sensing on the
worker-route validation problem.

In SimpleSim, we evaluate three route-repair scenarios
(\autoref{fig:scenario_results}). In the \textit{single-passage} scenario, the
nominal route crosses one passage that is traversable by the scout but not by the
worker, so the scout must discover a wider detour. In the \textit{double-passage}
scenario, the nominal route crosses two such passages. In the \textit{infeasible}
scenario, no complete worker-feasible route exists, so the correct behavior is to
terminate without returning an invalid worker route.

On the single- and double-passage scenarios, our method terminates after
\(169.8 \pm 59.1\)\,m and \(286.7 \pm 53.9\)\,m of scout travel, respectively. This
is \(37\%\) and \(24\%\) less than \textit{Worker-as-scout}, and \(71\%\) and \(52\%\)
less than \textit{Full exploration} (\autoref{fig:scenario_results}). On the
infeasible route, \textit{Worker-as-scout} travels only \(56.3\)\,m because its larger
footprint prevents it from entering the narrow passage and it stops before
inspecting the far region. In contrast, our scout passes through the narrow
region, attempts to reconnect the worker graph, and terminates after
eight failed reconnection attempts without validating a worker-feasible route. This requires \(145.1\)\,m of scout travel, which remains below
\textit{Full exploration}. 

\begin{figure}[t!]
\centering
\begin{subfigure}[b]{0.31\linewidth}
  \centering
  \includegraphics[width=\linewidth]{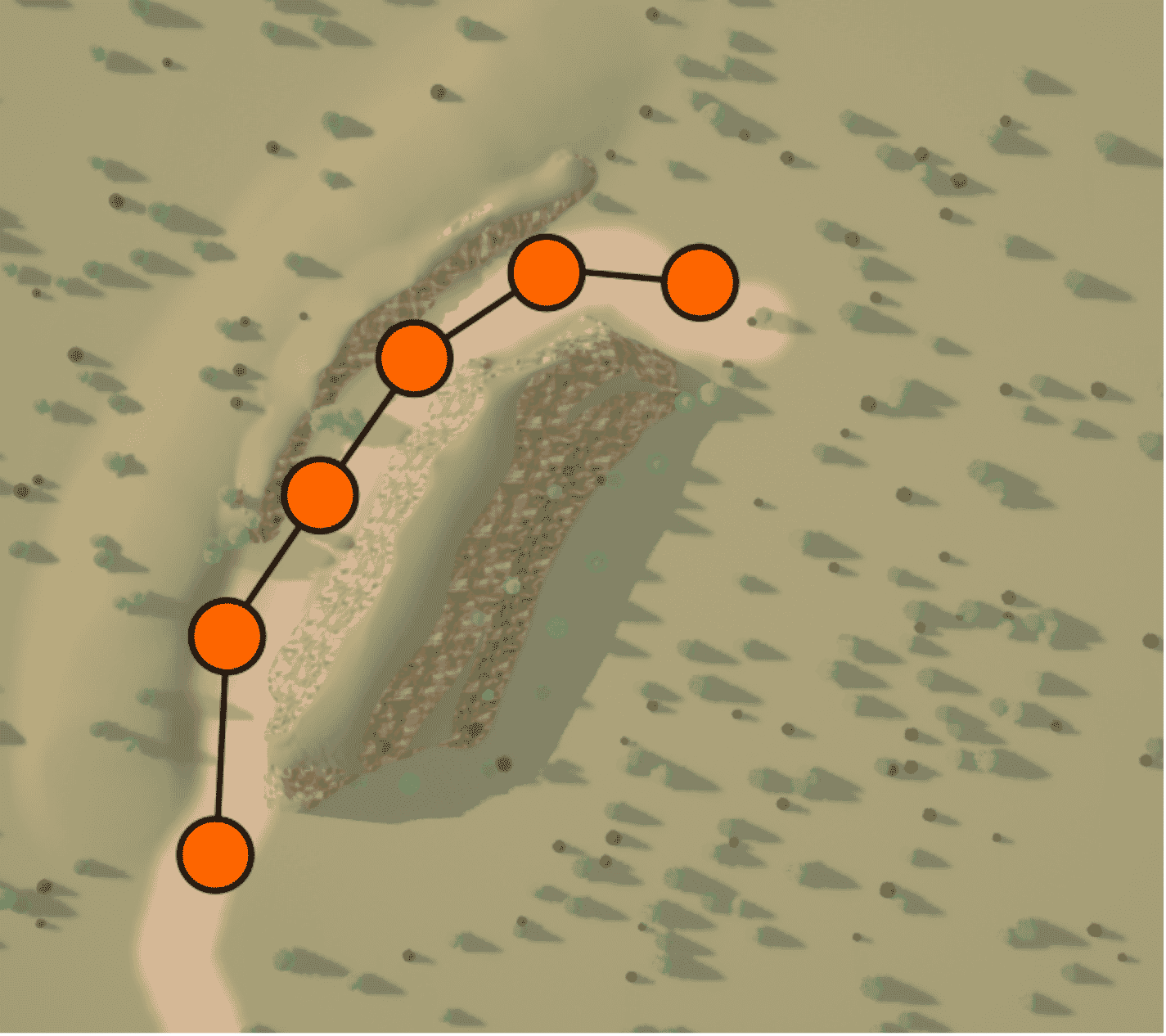}
  \caption{CARLA scenario.}
\end{subfigure}
\hfill
\begin{subfigure}[b]{0.31\linewidth}
  \centering
  \includegraphics[width=\linewidth]{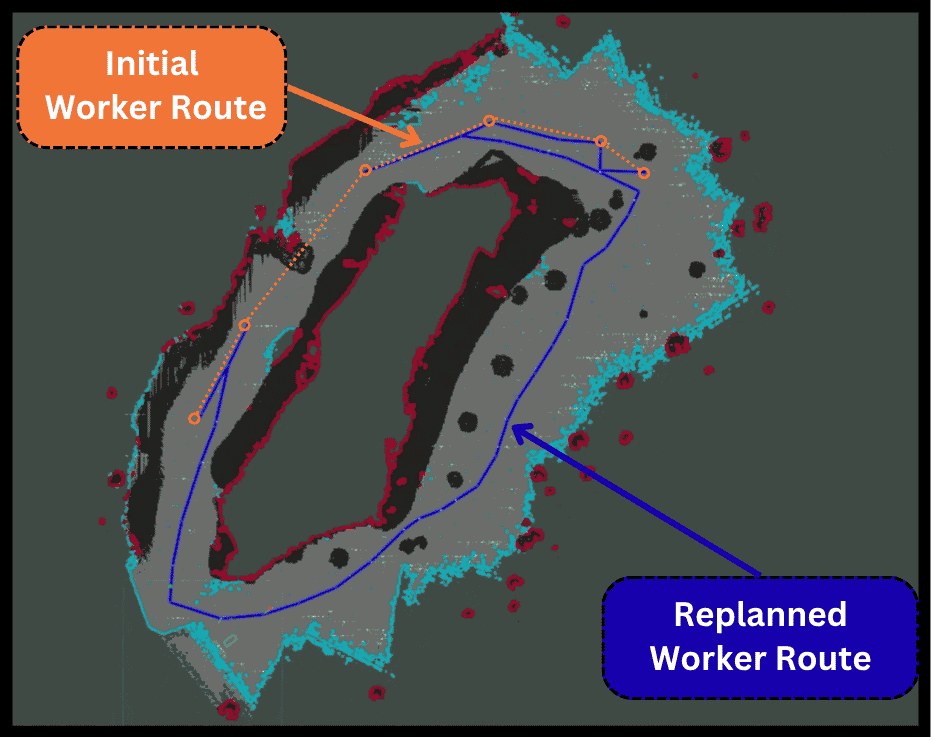}
  \caption{Explored area.}
\end{subfigure}
\hfill
\begin{subfigure}[b]{0.36\linewidth}
  \centering
  \includegraphics[width=\linewidth]{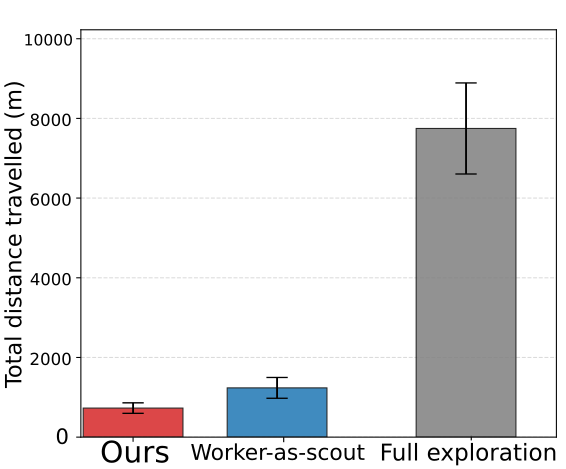}
  \caption{Scout travel distance \(D_{\text{term}}\).}
\end{subfigure}
\caption{CARLA results. (a) Representative scenario. (b) Explored map with the initial route and the worker-feasible replanned route returned by our method. (c) Total scout distance $D_{\text{term}}$ over five runs for \textit{Ours}, \textit{Worker-as-scout}, and \textit{Full exploration}.}
\label{fig:carla_results}
\vspace{-1.5em}
\end{figure}

We also evaluate the full framework in CARLA, where the method runs in a
\(365 \times 365\)\,m region with an ordered route through seven goals. The scout
is a \(4.85 \times 2.2\)\,m vehicle with a front-facing \(90^\circ\) depth camera,
and the worker is modeled as a larger \(6.0 \times 2.5\)\,m footprint. Under these
more realistic sensing and vehicle constraints, our method recovers a
worker-feasible route by exploring around the obstacle and repairing the initial
route where it is not worker-traversable (\autoref{fig:carla_results}). It travels
\(729.2 \pm 131.8\)\,m, which is \(41\%\) less than \textit{Worker-as-scout}
\((1236.1 \pm 261.3\)\,m) and \(91\%\) less than \textit{Full exploration}
\((7747.8 \pm 1142.8\)\,m).

\section{Real-World Demonstration}
\label{sec:real_life_experiments}
\begin{wrapfigure}{r}{0.44\textwidth}
    \centering
    \vspace{-4.5em}

    \begin{subfigure}[b]{0.34\linewidth}
        \centering
        \includegraphics[width=\linewidth]{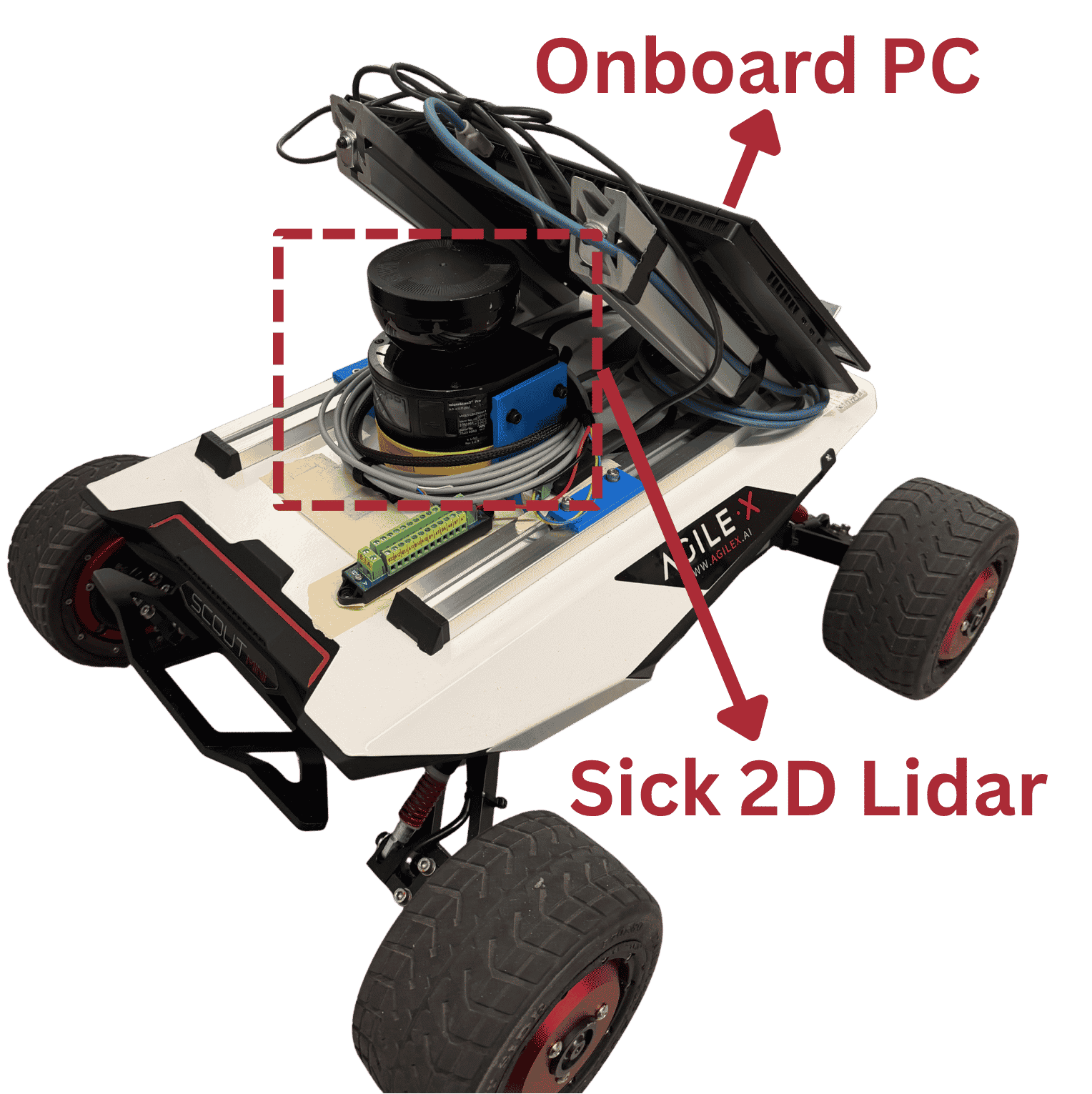}
        \caption{}
        \label{fig:rw_robot}
    \end{subfigure}
    \hfill
    \begin{subfigure}[b]{0.90\linewidth}
        \centering
        \includegraphics[width=\linewidth]{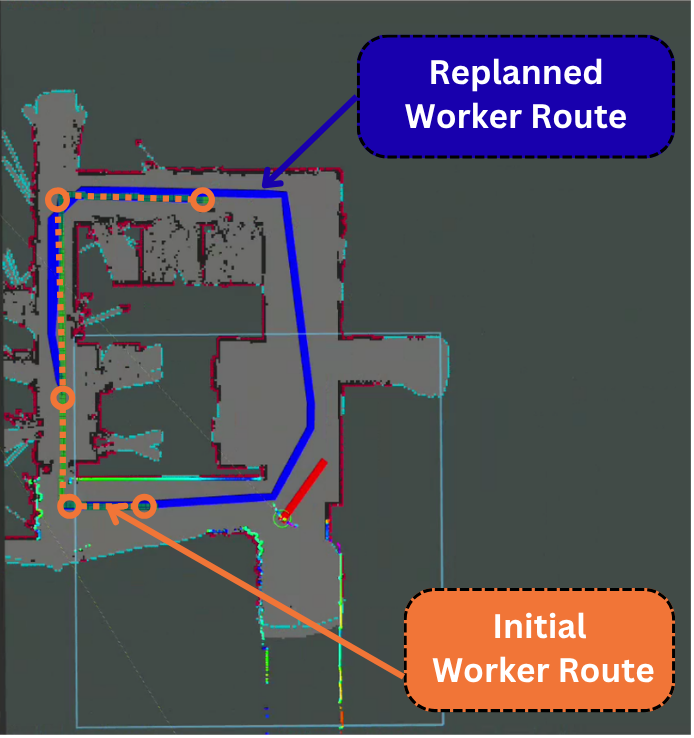}
        \caption{}
        \label{fig:rw_map}
    \end{subfigure}

    \caption{Real-world setup: (a) AgileX Scout Mini and (b) Explored area.}
    \label{fig:real_world_setup}
    \vspace{-2.0em}
\end{wrapfigure}
\begin{figure}[t]
    \centering
    \includegraphics[width=0.85\linewidth]{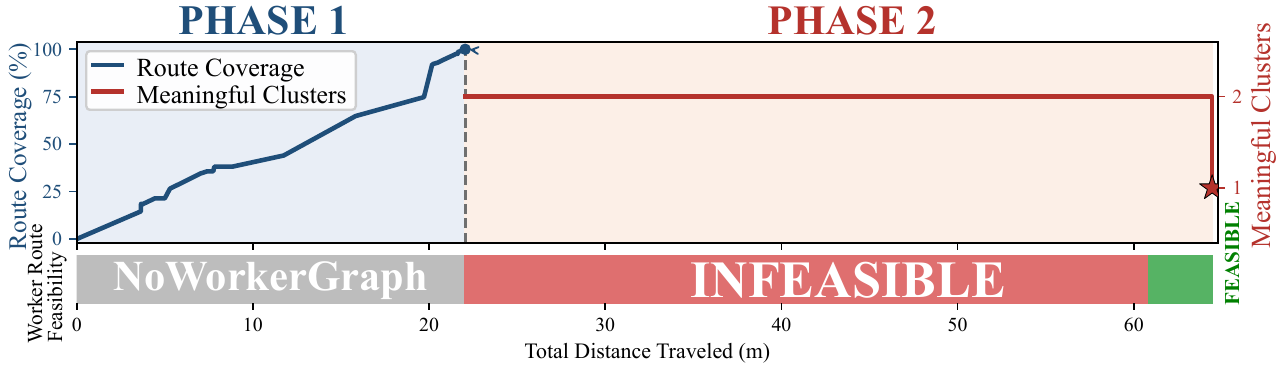}
    \caption{Two-phase progress during the real-world deployment.}
    \label{fig:rw_progress}
    \vspace{-1.5em}
\end{figure}

We validated the framework in the Unity Hall Building at Worcester Polytechnic
Institute using an AgileX Scout Mini equipped with a SICK 2D LiDAR
($180^\circ$ FOV, $5\,\mathrm{m}$ range) and an onboard PC, as shown in \autoref{fig:rw_robot}. The
experiment was conducted within a $50 \times 50\,\mathrm{m}$ planning
region represented by an occupancy grid with $0.1\,\mathrm{m}$
resolution. The worker platform was modeled with a footprint of
$1.5\,\mathrm{m} \times 0.95\,\mathrm{m}$, while the scout was modeled
as a circular robot with a $0.5\,\mathrm{m}$ footprint. 
As shown in
\autoref{fig:rw_map}, the environment contains a narrow passage that is
traversable by the scout but not necessarily by the larger worker,
making worker-route feasibility non-trivial.
\autoref{fig:rw_progress} summarizes the exploration progress.
Phase~1 achieved $100\%$ route coverage after $22.05\,\mathrm{m}$ of
travel. During the transition stage, the worker graph was found to
contain two distinct goal-containing components, triggering Phase~2. After
an additional $42.4\,\mathrm{m}$ of travel, the scout observed the
missing connection and merged the graph into a single connected
component, producing a replanned worker route of length
$17.76\,\mathrm{m}$ through all ordered goals.
The key insight is that complete route coverage alone does not guarantee worker-route feasibility. Even after the scout fully observed the preferred-route corridor, worker-graph connectivity was still required to validate a worker-feasible route. In \autoref{fig:rw_map}, the initial worker route (orange) crosses a narrow region that could not be validated for the larger worker footprint. Phase~1 completed observation of the route corridor, while Phase~2 targeted the unresolved worker-graph components and revealed the connection needed to validate the replanned route (blue). We used the same frontier-based viewpoint-selection pipeline with real LiDAR measurements, although we did not systematically evaluate robustness to sensing and localization uncertainties.
\vspace{-0.2em}
\section{Conclusion}
\label{sec:conclusion}
This paper addressed asymmetric scout--worker reconnaissance in unknown environments, where a small, agile scout must discover a worker-feasible path through ordered goal locations for a larger robot with stricter traversability constraints. The results show that the framework validates nominal worker routes, repairs blocked segments with observed feasible detours, and identifies cases where no complete worker-feasible route has been validated. Across SimpleSim and CARLA, it consistently reduced scout travel relative to baselines that either ignore scout--worker asymmetry or explore broadly, demonstrating the value of task-directed reconnaissance. A real-world indoor deployment on the second floor of WPI's Robotics Department further showed the practical character of the approach, with the scout navigating narrow corridors to discover a worker-feasible route. Overall, exploiting the scout's greater mobility while evaluating all returned paths under the worker's feasibility model enables route validation with substantially less exploration. Future work will extend the approach to non-nested, terrain-dependent traversability models, larger-scale environments, and physical scout--worker teams.

\bibliographystyle{spmpsci}
\bibliography{citations/refs}

@String { icra   = {{IEEE} Intl. Conf. Robot. Autom.} }

@String { ijrr   = {Intl. J. of Robotics Research} }

@String { iral   = {{IEEE} Robot. Autom. Letters} }

@String { tro    = {{IEEE} Trans. Robot.} }

@String { ssrr      = {{IEEE/RSJ} Symp. on Safety, Security, and Rescue Robotics} }

@String { bioeng  = {Biosystems Engineering} }

@String { scirobot= {Science Robotics} }

@String { jair = {J. Artif. Intell. Res.} }

@String { ai = {Artif. Intell.} }

@String { autorob = {Auton. Robots} }

@String { tssc     = {{IEEE} Trans. Syst. Sci. Cybern.} }

@String { jacm     = {J. ACM} }

@String { jmlr     = {J. Mach. Learn. Res.} }

@String { tevc     = {{IEEE} Trans. Evol. Comput.} }

@String { aaai = {{AAAI} Conf. on Artif. Intell.} }

@String { cira = {{IEEE} Intl. Symp. on Computational Intelligence in Robot. and Autom.} }

@article{vutetakis2025apn,
  author  = {Vutetakis, David and Xiao, Jing},
  title   = {Active Perception Network for Non-Myopic Online Exploration
             and Visual Surface Coverage},
  journal = ijrr,
  year    = {2025},
  volume  = {44},
  number  = {2},
  pages   = {247--272}
}

@article{rockenbauer2025traversing,
  author  = {Rockenbauer, Friedrich M. and Lim, Jaeyoung and
             M{\"u}ller, Marcus G. and Siegwart, Roland and Schmid, Lukas},
  title   = {Traversing {Mars}: {C}ooperative Informative Path Planning
             to Efficiently Navigate Unknown Scenes},
  journal = iral,
  year    = {2025},
  volume  = {10},
  number  = {2},
  pages   = {1776--1783}
}

@article{moon2025iatigris,
      author={Moon, Brady and Suvarna, Nayana and Jong, Andrew and Chatterjee, Satrajit and Yuan, Junbin and Cao, Muqing and Scherer, Sebastian},
      journal=tro, 
      title={IA-TIGRIS: An Incremental and Adaptive Sampling-Based Planner for Online Informative Path Planning}, 
      year={2026},
      volume={42},
      number={},
      pages={1695-1713}
}

@article{burusa2024semantic,
  author  = {Burusa, Akshay K. and Scholten, Joost and Wang, Xin and
             Rapado-Rinc{\'o}n, David and van Henten, Eldert J. and
             Kootstra, Gert},
  title   = {Semantics-Aware Next-Best-View Planning for Efficient Search
             and Detection of Task-Relevant Plant Parts},
  journal = bioeng,
  year    = {2024},
  volume  = {246},
  pages   = {248--262}
}

@inproceedings{kapoutsis2017cia,
  author    = {Kapoutsis, Athanasios Ch. and Malliou, Christina M. and
               Chatzichristofis, Savvas A. and Kosmatopoulos, Elias B.},
  title     = {Continuously Informed Heuristic {A*}-Optimal Path Retrieval
               Inside an Unknown Environment},
  booktitle = ssrr,
  year      = {2017},
  pages     = {221--227}
}

@article{cao2023tare,
  author  = {Cao, Chao and Zhu, Hongbiao and Ren, Zhongqiang and
             Choset, Howie and Zhang, Ji},
  title   = {Representation Granularity Enables Time-Efficient Autonomous
             Exploration in Large, Complex Worlds},
  journal = scirobot,
  year    = {2023},
  volume  = {8},
  number  = {80},
}

@inproceedings{kulkarni2022cohort,
  author    = {Kulkarni, Mihir and Dharmadhikari, Mihir and
               Tranzatto, Marco and Zimmermann, Samuel and
               Reijgwart, Victor and De Petris, Paolo and Nguyen, Huan and
               Khedekar, Nikhil and Papachristos, Christos and Ott, Lionel
               and Siegwart, Roland and Hutter, Marco and Alexis, Kostas},
  title     = {Autonomous Teamed Exploration of Subterranean Environments
               using Legged and Aerial Robots},
  booktitle = icra,
  year      = {2022},
  pages     = {3306--3313}
}

@article{naazare2022wgnbvp,
  author  = {Naazare, Menaka and Rosas, Francisco Garcia and Schulz, Dirk},
  title   = {Online Next-Best-View Planner for {3D}-Exploration and
             Inspection with a Mobile Manipulator Robot},
  journal = iral,
  year    = {2022},
  volume  = {7},
  number  = {2},
  pages   = {3779--3786}
}

@article{felner2004pha,
  author  = {Felner, Ariel and Stern, Roni and Ben-Yair, Asaph and
             Kraus, Sarit and Netanyahu, Nathan},
  title   = {{PHA*}: Finding the Shortest Path with {A*} in an Unknown
             Physical Environment},
  journal = jair,
  year    = {2004},
  volume  = {21},
  pages   = {631--670}
}

@article{peterson2018aerial,
  author  = {Peterson, John and Chaudhry, Haseeb and Abdelatty, Karim and
             Bird, John and Kochersberger, Kevin},
  title   = {Online Aerial Terrain Mapping for Ground Robot Navigation},
  journal = {Sensors},
  year    = {2018},
  volume  = {18},
  number  = {2},
  pages   = {630}
}

@inproceedings{bircher2016rhnbvp,
  author    = {Bircher, Andreas and Kamel, Mina and Alexis, Kostas and
               Oleynikova, Helen and Siegwart, Roland},
  title     = {Receding Horizon ``Next-Best-View'' Planner for
               {3D} Exploration},
  booktitle = icra,
  year      = {2016},
  pages     = {1462--1468}
}

@article{schmid2020online,
  author  = {Schmid, Lukas and Pantic, Michael and Khanna, Raghav and
             Ott, Lionel and Siegwart, Roland and Nieto, Juan},
  title   = {An Efficient Sampling-Based Method for Online Informative
             Path Planning in Unknown Environments},
  journal = iral,
  year    = {2020},
  volume  = {5},
  number  = {2},
  pages   = {1500--1507}
}

@inproceedings{koenig2002dlite,
  author    = {Koenig, Sven and Likhachev, Maxim},
  title     = {{D*} Lite},
  booktitle = aaai,
  year      = {2002}
}

@article{koenig2004lpa,
  author  = {Koenig, Sven and Likhachev, Maxim and Furcy, David},
  title   = {Lifelong Planning {A*}},
  journal = ai,
  year    = {2004},
  volume  = {155},
  number  = {1--2},
  pages   = {93--146}
}

@article{arora2019heterogeneous,
  author  = {Arora, Akash and Furlong, P. Michael and Fitch, Robert and
             Sukkarieh, Salah and Fong, Terrence},
  title   = {Multi-Modal Active Perception for Information Gathering in
             Science Missions},
  journal = autorob,
  year    = {2019},
  volume  = {43},
  number  = {7},
  pages   = {1827--1853}
}

@inproceedings{yamauchi1997frontier,
  author    = {Yamauchi, Brian},
  title     = {A Frontier-Based Approach for Autonomous Exploration},
  booktitle = cira,
  year      = {1997},
  pages     = {146--151}
}

@article{hart1968astar,
  author  = {Hart, Peter E. and Nilsson, Nils J. and Raphael, Bertram},
  title   = {A Formal Basis for the Heuristic Determination of Minimum
             Cost Paths},
  journal = tssc,
  year    = {1968},
  volume  = {4},
  number  = {2},
  pages   = {100--107}
}

@article{tarjan1975unionfind,
  author  = {Tarjan, Robert E.},
  title   = {Efficiency of a Good but not Linear Set Union Algorithm},
  journal = jacm,
  year    = {1975},
  volume  = {22},
  number  = {2},
  pages   = {215--225}
}

@article{DEAP_JMLR2012,
  author  = {F\'elix-Antoine Fortin and Fran\c{c}ois-Michel {De Rainville} and Marc-Andr\'e Gardner and Marc Parizeau and Christian Gagn\'e},
  title   = {{DEAP}: Evolutionary Algorithms Made Easy},
  journal = jmlr,
  year    = {2012},
  volume  = {13},
  pages   = {2171--2175},
  month   = jul
}

@article{deb2002nsga2,
  author  = {Deb, Kalyanmoy and Pratap, Amrit and Agarwal, Sameer and Meyarivan, T.},
  title   = {A Fast and Elitist Multiobjective Genetic Algorithm: {NSGA-II}},
  journal = tevc,
  year    = {2002},
  volume  = {6},
  number  = {2},
  pages   = {182--197}
}

@inproceedings{Dosovitskiy17,
  title = {{CARLA}: {An} Open Urban Driving Simulator},
  author = {Alexey Dosovitskiy and German Ros and Felipe Codevilla and Antonio Lopez and Vladlen Koltun},
  booktitle = {Proceedings of the 1st Annual Conference on Robot Learning},
  pages = {1--16},
  year = {2017}
}

\end{document}